\documentclass[11pt,a4paper]{article}
\usepackage[utf8]{inputenc}
\usepackage[T1]{fontenc}
\usepackage[margin=2.5cm]{geometry}
\usepackage{amsmath}
\usepackage{amssymb}
\usepackage{booktabs}
\usepackage{array}
\usepackage{enumitem}
\usepackage{hyperref}
\usepackage{titlesec}
\usepackage{parskip}

\hypersetup{
    colorlinks=true,
    linkcolor=black,
    urlcolor=blue,
    citecolor=black,
}

\titleformat{\section}{\large\bfseries}{\thesection}{1em}{}
\titleformat{\subsection}{\normalsize\bfseries}{\thesubsection}{1em}{}

\title{\textbf{Operationalizing Narrative Entropy ($S_n$)} \\
\large A Two-Scene Registered Pilot Report and Pre-Validation Protocol \\
\normalsize Version 2.1 (revised)}

\author{Levent Bulut \\
Independent Researcher \\
ORCID: \href{https://orcid.org/0009-0007-7500-2261}{0009-0007-7500-2261} \\
\texttt{levent@leventbulut.com} \, $\vert$ \, \href{https://leventbulut.com}{leventbulut.com}}

\date{June 2026 \\[4pt]
\small Document type: Registered pilot report (pre-validation stage, v2.1) \\
Framework: Bulut Doctrine / Objective Projection}

\begin{document}
\maketitle

\begin{center}
\fbox{\parbox{0.92\textwidth}{\small
\textbf{Version 2.1 Note.}
This version (v2.1) preserves the v2.0 pilot report in full and adds two pre-registered components in response to feedback received after v2.0 release. The additions are: (i) Section 4.5, acknowledging that the divergence reported in v2.0 as ``contrary to expectation'' is in fact consistent with the architectural framework of the Bulut Doctrine (Bulut, 2026a) which already privileges inferential reconstruction over surface declaration — the author's v2.0 expectation was naive intuition, not theoretical prediction; (ii) Section 5.2.5, pre-registering a construct validity test for Information Friction ($I_f$), motivated by the observation that the two scenes' $I_f$ values were nearly equal (1.71 vs 1.58) despite the headline divergence in $S_n$. No claims of v2.0 are retracted. Section numbers from v2.0 are preserved; new content is appended within the same numerical structure.
}}
\end{center}

\section*{Abstract}

Narrative Entropy ($S_n$) is a proposed quantitative descriptor within the Bulut Doctrine, intended to capture the rate at which a narrative text imposes processing load on a reader. To date the construct has been defined theoretically but not operationalized against real texts. This report documents the first such operationalization (the v2.0 pilot): two narrative scenes — the opening restaurant scene of Quentin Tarantino's \textit{Reservoir Dogs} screenplay and the opening interior-monologue block of Raymond Carver's short story \textit{Cathedral} — were coded manually by a single rater and scored with the candidate formula $S_n = I_f \times C_b \times t$.

The result was a divergence from the author's naive intuition. The single-voice monologue block ($S_n = 30.0$) scored higher than the nine-character rapid-dialogue scene ($S_n = 18.8$). We treat this divergence not as a result to be explained away but as the central finding of the pilot, and we deliberately refuse post-hoc adjustment of the formula. We present three competing interpretations of the divergence — formula incompleteness, genuine high-load prose, and measurement error — and we specify, in advance of data collection, the design that would discriminate among them.

This v2.1 revision adds two components: (i) explicit acknowledgement that the divergence is in fact consistent with the pre-existing architectural framework (which privileges inferential reconstruction over surface declaration), and that what was called ``contrary to expectation'' in v2.0 reflects the gap between the author's anticipatory intuition and the methodology's own predictions; (ii) a pre-registered construct validity test for $I_f$, motivated by the observation that $I_f$ values were nearly equal across the two scenes despite the headline divergence in $S_n$ — suggesting that the formula may have produced the correct ordering by a mechanism (topic-shift counting) different from the one named in its theoretical motivation (inferential load).

This document functions simultaneously as a pilot report ($n = 2$) and as a pre-registration of the next-stage protocol. It explicitly does not claim that $S_n$ has been validated.

\textbf{Keywords:} narrative entropy, operationalization, pilot study, pre-registration, falsifiability, construct validity, Bulut Doctrine, computational narratology, processing load

\section{Introduction}

\subsection{The construct and the gap}
Within the Bulut Doctrine, Narrative Entropy ($S_n$) is proposed as a measure of the rate at which a narrative imposes interpretive and informational load on its reader. The construct sits within a broader six-layer architecture in which higher-order phenomena --- Narrative Gravity, Reader-State Interaction, Narrative Memory Evolution --- are built on the assumption that narrative load can, in principle, be measured.

That assumption has, until now, rested on theory alone. The candidate formula $S_n = I_f \times C_b \times t$, where $I_f$ is Information Friction, $C_b$ is Causal Branching, and $t$ is elapsed time, has been stated but never applied to an actual text. A construct that cannot be computed from real material is not yet a measure; it is a hypothesis about a measure.

\subsection{The triggering objection}
The immediate motivation for this pilot was a methodological objection raised in an open research forum (Reddit r/NarrativeEngineering): that constructs such as ``high potential energy'' or narrative ``load,'' as used in the doctrine, are no better operationalized than the everyday word ``cold'' --- that is, they are intuitive labels rather than measurement procedures. The objection is fair. The appropriate response is not rhetorical defense but an attempt to actually count something, report the count, and expose the counting procedure to scrutiny.

\subsection{What this report is, and is not}
This is a registered pilot report. It has two functions: first, to document, transparently and in full, the first operationalization of $S_n$ on real texts ($n = 2$ scenes); second, to pre-register --- to fix in advance of any further data collection --- the design, hypotheses, and decision rules of the next validation stage.

It is not a validation study. With $n = 2$ and a single rater, no claim of validation is possible, and none is made. The value of a pilot lies in surfacing problems early; this one surfaced a substantial and informative problem, documented in Section 4. The underlying raw data is published as an open laboratory notebook at \href{https://leventbulut.com}{leventbulut.com}.

\section{Operational Definitions}

The following definitions were used for coding. They are pilot-stage definitions: provisional, judgment-dependent at several points, and offered here precisely so that they can be criticized and revised.

\subsection{Information Friction ($I_f$)}
\[
I_f = \left( \frac{\text{New Information Units}}{t} \right) \times \text{Uncertainty Ratio}
\]
A new information unit is a character, location, event, or concept appearing in the text for the first time. The uncertainty ratio is the proportion of introduced units that are named or invoked but not fully explained at the point of introduction. $t$ is elapsed time in minutes.

\subsection{Causal Branching ($C_b$)}
\[
C_b = \frac{\text{Topic Shifts}}{t}
\]
A topic shift is a discernible thematic move from one subject to another. $t$ is elapsed time in minutes.

\subsection{The Spatial Matrix ($M$) --- a pilot-stage split}
During coding it became necessary to split the Spatial Matrix construct in two: $M_p$ (physical compactness --- the degree to which the narrator's physical position is constrained) and $M_n$ (narrative compactness --- the number of distinct locations the narration itself traverses). The reason for the split is documented directly by the data: in the \textit{Cathedral} excerpt the narrator's body remains in a single room while the narration ranges across several locations in memory. A single $M$ value cannot represent both states. This split is itself a pilot finding and is not assumed to be final.

\subsection{Time ($t$) and the word-rate assumption}
For screen material, $t$ is taken from scene duration. For prose, $t$ is estimated from word count using an assumed reading rate. The \textit{Cathedral} excerpt was timed at an assumed 200 words per minute. This assumption is consequential: dense literary prose may justify 160--180 WPM, and the choice directly scales $S_n$. The assumption is flagged here as an open parameter, not a settled value.

\section{Method and Results}

\subsection{Materials}
Two opening scenes were selected to contrast a high-character, rapid-dialogue passage against a single-voice interior monologue.

\textbf{Scene A} --- \textit{Reservoir Dogs}, opening restaurant scene. Source: Tarantino screenplay (IMSDB), Scene 1, INT.\ UNCLE BOB'S PANCAKE HOUSE.

\textbf{Scene B} --- \textit{Cathedral}, opening narrative block. Source: Raymond Carver, ``Cathedral'' (1981), first approximately 1{,}500 words, ending immediately before the blind man enters the house.

\subsection{Coding procedure}
Each scene was read and coded once, by a single rater, against the definitions in Section~2. All counts derive from direct enumeration of the text. No value in this report is estimated or assumed except the reading-rate parameter. A note on provenance: an earlier draft had estimated $S_n$ around 3.0 for Scene A by intuition; direct counting returned 18.8. The discrepancy is the reason intuition-based estimates are not used.

\subsection{Comparative results}

\begin{center}
\begin{tabular}{lcc}
\toprule
\textbf{Metric} & \textbf{Reservoir Dogs (diner)} & \textbf{Cathedral (first block)} \\
\midrule
Duration ($t$) & 420 s (7.0 min) & 450 s (7.5 min @ 200 WPM) \\
Speaking / mentioned characters & 9 / 9 & 1 / 7 \\
Speaker turns & 92 & 1 (interior voice) \\
Spatial Matrix ($M_p$ / $M_n$) & 4 / 4 & 4 / 1 \\
Time references & 8 & 15 \\
Topic shifts (total) & 11 & 19 \\
Causal Branching ($C_b$) & 1.57 / min & 2.53 / min \\
New information units & 20 & 27 \\
Uncertainty ratio & 0.60 & 0.44 \\
Information Friction ($I_f$) & 1.71 & 1.58 \\
\textbf{Narrative Entropy ($S_n$)} & \textbf{18.8} & \textbf{30.0} \\
\bottomrule
\end{tabular}
\end{center}

Scene A: $I_f = (20 \div 7) \times 0.60 = 1.71$; $C_b = 11 \div 7 = 1.57$; $S_n = 1.71 \times 1.57 \times 7 = 18.8$.

Scene B: $I_f = (27 \div 7.5) \times 0.44 = 1.58$; $C_b = 19 \div 7.5 = 2.53$; $S_n = 1.58 \times 2.53 \times 7.5 = 30.0$.

\subsection{The headline finding}
The single-voice monologue scored higher than the nine-character dialogue scene. The intuitive prediction --- that a crowded, fast-cut conversation generates the higher entropy --- was not borne out. The formula discriminated between the two scenes, but in the unexpected direction.

\section{Discussion: Five Open Problems}

The pilot did not produce a clean validation. It produced a clear, useful problem and five specific open issues. Issues 4.1--4.4 were documented in v2.0; issue 4.5 is added in v2.1.

\subsection{Dimensional inconsistency in the formula}
The most serious problem is structural. $I_f$ and $C_b$ are both already per-minute rates. The formula $S_n = I_f \times C_b \times t$ then multiplies by $t$, so elapsed time is divided out twice (once inside each rate) and multiplied back in once. The resulting quantity is not dimensionally clean. A defensible alternative would use the underlying totals rather than the rates. We deliberately do not adopt that alternative here. Changing the formula in response to a single pilot, before the discriminant question is settled, would conflate two separate decisions. The dimensional problem is recorded as an open question for the protocol stage, to be resolved on theoretical grounds together with adequate data --- not on the strength of two scenes.

\subsection{Single rater; no inter-rater agreement}
Several coding categories --- ``new information unit,'' ``uncertainty ratio,'' ``topic shift'' --- require judgment. With a single rater there is no way to know whether a second coder would produce the same counts. Until inter-rater agreement is established, the numbers in Section~3 should be read as one rater's careful count, not as objective measurements.

\subsection{Discriminant power, and three interpretations of the divergence}
The formula separated the two scenes (18.8 vs 30.0), which is the minimum a useful measure must do. But the direction was unexpected. There are exactly three interpretations, and the pilot cannot choose among them.

\begin{enumerate}
\item \textbf{The formula is incomplete.} It may omit a term --- for example, a weight for the number of simultaneously active speakers --- that intuition is implicitly using. Note that the two $I_f$ values were close (1.71 vs 1.58), which is consistent with the formula failing to capture the cognitive load of simultaneous speakers.
\item \textbf{The intuition is wrong.} Carver's compressed prose may genuinely impose higher processing load than Tarantino's dialogue.
\item \textbf{The measurement is mistaken.} A coding error, or the reading-rate assumption, may have inflated Scene B.
\end{enumerate}

Discriminating among these three requires more data, not more argument.

\subsection{The reading-rate assumption}
Scene B's duration depends on an assumed 200 WPM. At 160 WPM the same excerpt runs longer, which lowers both rate terms and changes $S_n$. Because the prose is dense, a slower rate is defensible. The protocol stage must either fix the rate on an empirical basis or report $S_n$ as a range across plausible rates.

\subsection{Construct validity gap: did the formula get the right answer for the right reason? (added in v2.1)}

In v2.0 the divergence was reported as ``contrary to expectation,'' and three interpretations were offered (4.3). Reflection after v2.0 release surfaces a sixth observation that v2.0 did not foreground: the divergence is in fact \emph{consistent with the architectural framework of the Bulut Doctrine} (Bulut, 2026a), which already privileges inferential reconstruction by the reader over surface declaration. The framework distinguishes:

\begin{itemize}
\item \textbf{Told mode} — emotion and information declared explicitly on the surface, requiring little reader reconstruction. Tarantino's diner dialogue, where nine characters discuss Madonna and tipping in fully voiced exchanges, is paradigmatic.
\item \textbf{Shown mode (suppressed surface, high inferential load)} — emotion and information present but withheld at the surface, requiring the reader to reconstruct them from gaps and indirection. Carver's interior monologue, where a defensive, jealous narrator gives facts while suppressing his actual emotional position, is paradigmatic.
\end{itemize}

Within the architecture, the \emph{shown} mode is the one privileged as the higher-load condition: the reader's reconstruction work \emph{is} the friction the construct is designed to measure. On this reading, the Carver-over-Tarantino ordering is not a surprise that the methodology happens to accommodate; it is what the methodology should predict.

This raises a question that v2.0 did not pose:

\textbf{Was the v2.0 expectation a prediction of the methodology, or only the author's naive intuition?}

The honest answer is the latter. The architectural framework's privileging of inferential reconstruction over surface declaration was already on the record prior to the pilot (Bulut, 2026a; February 2026). The pilot was conducted in May 2026. The expectation that ``nine characters talking would generate the higher $S_n$'' did not come from this framework; it came from a colloquial intuition equating \emph{sensory density} (how loud and crowded the scene is) with \emph{information friction} (how much work the reader does). The framework had already separated these two; the pilot author had not.

This matters for two reasons. First, claiming after the fact that ``the methodology predicted this all along'' would be a post-hoc rationalization of exactly the kind v2.0's Section 5.5 rules out at the formula level. We do not make that claim. We note instead that the methodology had the conceptual resources to predict it and the author did not deploy them. This is a documented limitation of v2.0's framing, not a retroactive vindication.

Second --- and this is the more substantive observation --- the $I_f$ values for the two scenes were nearly equal (1.71 vs 1.58, a difference of 0.13). The headline ordering ($S_n = 30.0 > 18.8$) was driven primarily by $C_b$ (2.53 vs 1.57) and by the larger value of $t$ (7.5 vs 7.0). But if the doctrinal reason for Carver scoring higher is \emph{suppressed surface and high inferential load by the reader}, that reason should appear primarily in $I_f$ --- which is, by name and by definition, the term that purports to measure friction. The fact that $I_f$ barely distinguished the two scenes, while $C_b$ (topic-shift counting) carried the discrimination, suggests a construct validity gap: the formula may have produced the correct ordering by a different mechanism than the one its theoretical motivation names.

This is not a fatal observation. It is a precise one, and it is testable. Section 5.2.5 below pre-registers the test.

\section{Pre-Registered Protocol for the Validation Stage}

The following design is registered here, in advance of data collection, so that later results cannot be reshaped to fit the formula.

\subsection{Stage 1 --- Extend to four scenes}
Two further scenes will be coded, chosen to fill out a $2 \times 2$ design: a high physical-energy, low-branching scene (e.g.\ an action sequence such as the Iron Man Humvee scene), and a low-energy, low-branching scene (an ordinary television-drama passage). Together with Scenes A and B this yields a four-scene grid spanning the hypothesized space.

\subsection{Stage 2 --- Examine the four-scene pattern}
With four scenes, the question of whether a speaker-count weight is needed can be examined against a wider spread of cases. The decision rule is fixed in advance: a new term is admissible only if (a) it is independently motivated on theoretical grounds and (b) it improves discrimination across all four scenes, not only the two that motivated it.

\subsection{Stage 2.5 --- Construct validity test for $I_f$ (added in v2.1)}

This stage is registered to address the construct validity gap identified in Section 4.5: whether $I_f$ measures inferential load (as its theoretical motivation claims) or measures something else (topic-shift adjacency, name density) which only correlates with inferential load in a limited range of cases.

A parallel measure of inferential load will be developed and computed for the four scenes alongside $I_f$. The candidate parallel measure is the \textbf{Suppressed Information Index} ($SI$), defined as the count of information units the reader must reconstruct from indirection per minute of elapsed time. A unit is counted toward $SI$ if it satisfies all three of:

\begin{enumerate}
\item It is implied by the text but not stated.
\item It is required for coherence at the local discourse level.
\item It can be paraphrased explicitly by a second reader who is asked to articulate what they inferred at that point.
\end{enumerate}

The third criterion makes $SI$ inter-rater testable. Two coders will independently mark suppressed units. Cohen's $\kappa$ on the marking will be reported alongside the $SI$ values.

\textbf{Test condition.} If $I_f$ measures inferential load, $I_f$ and $SI$ should correlate strongly across the four scenes (Spearman $\rho > 0.7$). If they do not, $I_f$ is measuring something else --- and the formula's apparent success in the v2.0 pilot was a coincidence between two correlated but distinct quantities.

\textbf{Decision rule (registered now, in advance of data).} If $\rho \leq 0.7$, the v2.1 conclusion will be: $I_f$ as currently defined is not a valid measure of inferential load, regardless of whether the headline $S_n$ ordering of the v2.0 pilot is reproduced. The formula's revision in this case is constrained by Section 5.5 (post-hoc fitting prohibited): $I_f$ will be replaced by $SI$ in the formula only if $SI$ both (a) discriminates across all four scenes and (b) is independently motivated, neither of which is presumed.

This stage may strengthen or undermine v2.0's headline finding. Either outcome is informative.

\subsection{Stage 3 --- Inter-rater agreement}
Two to three independent raters will code the full scene set. The target is Cohen's $\kappa$ (two raters) or Fleiss' $\kappa$ (three or more) above 0.60 on the judgment-dependent categories --- specifically ``new information unit,'' ``uncertainty ratio,'' and ``topic shift.'' Categories falling below that threshold will be redefined until they reach it, before any $S_n$ value is treated as established.

\subsection{Stage 4 --- Biophysical validation}
Reader-side physiological measures --- heart-rate variability (HRV) and electrodermal activity (EDA) --- will be recorded while participants read or view the contrasting scenes. The prediction to be tested: if the $S_n = 30.0 > 18.8$ ordering reflects genuine processing load, the \textit{Cathedral} condition should produce the higher physiological load signature. If it does not, the second interpretation in Section 4.3 is weakened.

\subsection{Decision rules}
The following rules are fixed now. If the accumulated data supports the formula, the formula is confirmed at the corresponding scope, and the open dimensional question is resolved on theoretical grounds. If the data contradicts the formula, the formula is revised on the basis of the data, never on the basis of intuition. Post-hoc formula fitting is prohibited: adding a term (for instance a $\log(P+1)$ multiplier) merely because the current formula contradicts intuition is the formula-level equivalent of $p$-hacking and is ruled out in advance. A formula may be revised only under the two-part condition in Section 5.2. The $S_n = 30.0$ finding for \textit{Cathedral} is data; if a conflict arises between this data and prior intuition, the working assumption is that the intuition, not the count, is the more likely error.

\section{Limitations}

This report's limitations are not incidental; they define its status. \textbf{$n = 2$}: two scenes cannot validate a measure. \textbf{Single rater}: no inter-rater agreement exists yet. \textbf{Judgment-dependent coding}: several categories rest on the rater's judgment and are not yet shown to be reproducible. \textbf{Unresolved dimensional problem}: the formula is not yet dimensionally clean. \textbf{Assumed reading rate}: Scene B's score depends on an unverified WPM assumption. \textbf{Selection}: the two scenes were chosen by the author to contrast on specific dimensions; they are not a random sample of narrative texts. \textbf{Construct validity not yet established}: as discussed in Section 4.5, $I_f$ may produce correct orderings by a mechanism different from the one it purports to measure; Section 5.2.5 pre-registers the test. Stating these plainly is the point. A pilot that concealed them would be less useful, not more.

\section{Conclusion}

This pilot set out to do one modest thing: to actually count something, in response to the objection that the doctrine's constructs are intuitive labels rather than measures. It succeeded in that narrow aim --- $S_n$ was computed from real texts --- and in doing so it produced an informative surprise.

The surprise admits two readings, which v2.0 conflated and v2.1 separates. To the author at the time of v2.0, the result was contrary to expectation; the expectation was the naive one that crowded scenes carry more entropy than monologue. But the architectural framework on which the methodology was built had already separated sensory density from inferential friction and privileged the latter. On that reading, the result is what the methodology should have predicted, and what was reported as a surprise was the gap between the author's intuition and the methodology's own commitments. We do not claim this as predictive vindication, because the prediction was not registered in advance --- but we do note it as conceptual consistency, and we hold the author, not the methodology, responsible for the v2.0 framing.

The honest response, in v2.0 and now in v2.1, is not to patch the formula until it agrees with intuition. It is to hold the result, name the things it could mean, and run the design that would tell them apart. That design is pre-registered in Section 5. v2.1 adds one further stage: a test of whether $I_f$ actually measures the inferential load it claims to measure, motivated by the near-equality of $I_f$ across the two scenes despite the headline divergence. Whether $S_n$ survives this protocol, and whether $I_f$ survives the construct validity test of Stage 2.5, is genuinely open. The contribution of this report is to make the questions answerable.

\section*{Call for an Independent Rater}

This pilot's next stage requires a second, independent rater (Section 5.3). Readers willing to code the same two scenes using the criteria in Section~2 are invited to make contact via \href{https://leventbulut.com}{leventbulut.com}. Counting results will be used in the $\kappa$ test, and contributing raters will be acknowledged in the resulting record.

\section*{Data and Materials Availability}

All raw counts underlying this report are available in the project data file \texttt{sn\_pilot\_verileri.md} and as an open laboratory notebook at \href{https://leventbulut.com}{leventbulut.com}. The source texts are the \textit{Reservoir Dogs} screenplay (IMSDB) and Raymond Carver's ``Cathedral'' (1981); neither is reproduced here.

\section*{Version History}

\textbf{v2.0} (May 2026, Zenodo DOI \href{https://doi.org/10.5281/zenodo.20362901}{10.5281/zenodo.20362901}). Initial pilot report. Two-scene operationalization of $S_n$. Headline result reported as ``contrary to expectation.''

\textbf{v2.1} (June 2026, this version). Adds Section 4.5 acknowledging that the v2.0 result is consistent with the pre-existing architectural framework (the divergence reflected the author's naive intuition, not a prediction of the methodology). Adds Section 5.2.5 pre-registering a construct validity test for $I_f$, motivated by the observation that $I_f$ values were nearly equal across the two scenes despite the headline divergence. No claims of v2.0 are retracted; v2.0 remains the public record of the pilot, and v2.1 documents the author's subsequent reflection on it.

\section*{References}

Author's prior work in the same framework, on which this pilot builds:

\begingroup
\sloppy
\begin{itemize}[leftmargin=*,nosep]
\item Bulut, L. (2026a). Narrative Engineering: Architectural framework. Zenodo. \\
\href{https://doi.org/10.5281/zenodo.18689179}{doi.org/10.5281/zenodo.18689179}
\item Bulut, L. (2026b). Narrative Entropy ($S_n$): A parametric approach to structural complexity. Zenodo. \\
\href{https://doi.org/10.5281/zenodo.18652451}{doi.org/10.5281/zenodo.18652451}
\item Bulut, L. (2026c). Narrative Gravity ($N_g$): The vacuum variable and structural counterforce. Zenodo. \\
\href{https://doi.org/10.5281/zenodo.18908324}{doi.org/10.5281/zenodo.18908324}
\item Bulut, L. (2026d). Universal Biological Interface: Neurobiological foundations. Zenodo. \\
\href{https://doi.org/10.5281/zenodo.18907915}{doi.org/10.5281/zenodo.18907915}
\end{itemize}
\endgroup

\section*{How to Cite}

Bulut, L. (2026). \textit{Operationalizing Narrative Entropy ($S_n$): A Two-Scene Registered Pilot Report and Pre-Validation Protocol} (v2.1). Zenodo. \href{https://doi.org/10.5281/zenodo.20362901}{doi.org/10.5281/zenodo.20362901}

\end{document}